\documentclass[review]{elsarticle}

\graphicspath{{./figures/}}

\usepackage{hyperref}
\usepackage{float}
\usepackage{verbatim}
\usepackage{apalike}
\usepackage{amssymb}
\usepackage{graphicx}
\restylefloat{figure}
\floatstyle{plaintop}
\restylefloat{table}
\usepackage{algorithm}
\usepackage{algpseudocode}
\usepackage{amsmath}

\makeatletter
\def\ps@pprintTitle{%
    \let\@oddhead\@empty
    \let\@evenhead\@empty
    \def\@oddfoot{}%
    \let\@evenfoot\@oddfoot
}
\makeatother

\biboptions{authoryear}

\begin{document}

\begin{frontmatter}

\title{CredWise: A Controlled Agentic Decision-Intelligence Framework for Explainable and Auditable Credit-Risk Assessment}

\author[label1]{Aakash Kumar Tiwari\corref{cor1}}
\ead{tiwariaakash1025@kgpian.iitkgp.ac.in}

\cortext[cor1]{Corresponding author.}

\address[label1]{Department of Mathematics, Indian Institute of Technology Kharagpur, Kharagpur 721302, West Bengal, India}


\begin{abstract}
Credit-risk prediction is important in banking, but a prediction alone does not explain why an applicant is risky or how it should be combined with other evidence. This paper presents CredWise, a decision-support framework that integrates credit-risk prediction, probability calibration, explainable artificial intelligence, policy retrieval, SQL analytics, and controlled agent-based workflows. An XGBoost model is trained on Lending Club data (1,345,310 loans, 18 features) using a temporal split: 2007--2016 for training, 2017 for validation, and 2018 for testing. On the 2018 test set, the calibrated model achieved a ROC-AUC of 0.7109, PR-AUC of 0.2993, F1-score of 0.3714, and accuracy of 65.44\%. Calibration reduced the Brier score from 0.2157 to 0.1273 and the expected calibration error from 0.2862 to 0.0585. SHAP explanations were temporally stable, with a Spearman correlation of 0.9959 between 2017 and 2018 feature rankings. On 28 labeled queries covering nine policy sections, FAISS achieved the best Hit@1 (0.929) and MRR (0.964), while all three retrieval methods reached Hit@5 = 1.0. Agent routing achieved 95.6\% accuracy (43 of 45 cases), and the SQL benchmark scored 1.0 on exact-match, execution-success, and result-match across six cases. These results show that CredWise can combine predictions, explanations, policy evidence, and structured analytics in one controlled workflow. It is an academic research prototype, and final decisions remain with a human reviewer.
\end{abstract}


\begin{keyword}
credit risk \sep
explainable artificial intelligence \sep
probability calibration \sep
retrieval-augmented generation \sep
agentic artificial intelligence \sep
decision intelligence \sep
human-in-the-loop 
\end{keyword}
\end{frontmatter}

\section{Introduction}
Credit risk assessment is an important task in the financial sector. Banks and lending institutions need to estimate whether a borrower is likely to repay a loan. Traditional credit scoring has mainly used statistical classification methods, while machine learning has been adopted to capture more complex patterns in credit data \citep{hand1997statistical,lessmann2015benchmarking}. However, the use of machine learning in credit risk also raises practical issues, including how to interpret model predictions, how reliable the predicted probabilities are, and how the predictions can be used in an actual decision-support process \citep{bussmann2021explainable,advancingcredit2024}.
A credit-risk model usually produces a probability of default, but this probability alone may not be enough for a lending decision. A reviewer may also want to understand why an applicant was given a high-risk score, which features affected the prediction, and whether the result is consistent with the lending policy. This is particularly relevant for models such as gradient boosting, which can learn complex relationships that are not easy to interpret directly \citep{chen2016xgboost,friedman2001gradient}. Explainable artificial intelligence (XAI) methods can provide additional information about these predictions \citep{arrieta2020xai}, and explainability is now an important part of reviewing machine-learning-based credit decisions \citep{bussmann2021explainable,bucker2022lending}.
Another issue is whether the predicted probabilities are reliable. A model can distinguish between higher-risk and lower-risk applicants reasonably well, while the predicted probabilities may still differ from the actual outcome frequencies. Probability calibration is used to address this issue \citep{niculescu2005predicting,platt1999probabilistic,guo2017calibration}, and the Brier score is commonly used to measure the quality of probabilistic predictions \citep{brier1950verification}. Therefore, a credit-risk decision-support system should consider both discrimination and probability quality instead of relying only on classification performance.
The way a model is evaluated is also important because borrower characteristics, lending practices, and economic conditions can change over time. This is related to concept drift, where the relationship between the input data and the target variable changes over time \citep{gama2014conceptdrift}. A random train-test split may not capture such changes. Using earlier data for training and later data for testing can provide a more realistic view of how the model behaves on a later period.
A practical decision-support system may also require information outside the prediction model, such as lending policy documents. Retrieval-Augmented Generation (RAG) combines information retrieval with language-model based processing \citep{lewis2020rag}. Dense retrieval and sentence embeddings can be used to represent text for retrieval \citep{karpukhin2020dpr,reimers2019sbert}, while libraries such as FAISS support efficient similarity search \citep{johnson2019faiss}. In a credit-risk setting, this can be used to retrieve relevant policy evidence and present it together with a model prediction. Recent agent-based systems also allow language models to use tools and perform multi-step tasks \citep{yao2023react,wang2023agentsurvey}. However, a financial decision-support system needs controlled routing of tasks, restricted access to data and tools, and a clear distinction between model evidence, retrieved policy information, and generated text. Human oversight is also important because automated systems should assist users rather than remove human responsibility from important decisions \citep{amershi2019guidelines}. Earlier work studied multi-agent and retrieval-based workflows for enterprise customer support \citep{tiwari2026shopease}. The present work applies related ideas to credit-risk decision support, where predictions, policy evidence, explanations, and structured analytics need to be brought together.
Based on these requirements, this paper presents \textit{CredWise}, an academic research prototype for controlled credit-risk decision support. The system uses an XGBoost classifier and calibrates its output probabilities before they are used. It uses SHAP for feature-level explanations of individual predictions \citep{lundberg2017shap}, a policy retrieval component over a lending-policy document, a SQL analytics component for structured portfolio analysis, and a controlled agent workflow to connect these components. The system is evaluated on a large Lending Club dataset using a temporal split, with 2007--2016 used for training, 2017 for validation, and 2018 kept as an untouched test period. The final processed dataset contains 1,345,310 loan records and 18 prediction features. The evaluation covers classification performance, probability calibration, temporal stability of explanations, feature drift, error and subgroup analysis, policy retrieval, SQL analytics, and agent routing. Bootstrap confidence intervals are also reported for the final test-set metrics \citep{efron1986bootstrap}.
CredWise is not designed to replace a human lending decision. Instead, it brings different types of evidence into one decision-support workflow. The model prediction, explanation, retrieved policy evidence, and supporting analytics remain separate and traceable parts of the process, while the final decision is left to a human reviewer.
The main contributions of this work are as follows:
\begin{itemize}
    \item We develop a controlled credit-risk decision-support framework that combines prediction, probability calibration, explainability, policy retrieval, SQL analytics, and agent-based workflow control.
    
    \item We evaluate the credit-risk model using a temporal train-validation-test setup (2007--2016 / 2017 / 2018).
    
    \item We study the effect of probability calibration and report both discrimination and calibration metrics for the final test period.
    
    \item We evaluate the temporal stability of SHAP-based feature importance and examine feature drift over time.
    
    \item We evaluate the retrieval, SQL, and agent-routing components using separate controlled test cases instead of treating the complete system as a single black box.
    
    \item We keep the system human-in-the-loop, with model and retrieval outputs provided as decision-support evidence and the final decision left to the reviewer.
\end{itemize}
The remainder of this paper is organized as follows. Section~2 discusses related work. Section~3 presents the research questions. Section~4 describes the dataset and experimental setup. Section~5 presents the CredWise framework. Section~6 reports the experimental results. Section~7 discusses the findings. Section~8 describes the limitations. Section~9 concludes the paper.
\section{Related Work}
The work related to CredWise covers four main areas: credit-risk modelling, explainability and probability calibration, retrieval and agent-based systems, and human-in-the-loop decision support.

\subsection{Credit-Risk Modelling}
Credit scoring has traditionally used statistical classification methods, as discussed by Hand and Henley \citep{hand1997statistical}. With the use of machine learning, many classification algorithms have also been applied to credit scoring. Lessmann et al. \citep{lessmann2015benchmarking} compared a large number of these algorithms and showed that machine learning can provide useful predictive performance for credit scoring. More recent research has also discussed practical issues related to the use of machine learning for credit risk in financial applications \citep{advancingcredit2024}. Tree-based boosting methods are well suited to structured tabular data. XGBoost \citep{chen2016xgboost} provides an efficient implementation of gradient tree boosting \citep{friedman2001gradient} and is widely used for classification problems.
These studies mainly focus on prediction performance. A decision-support system, however, may also need to show how reliable a predicted probability is, why a particular prediction was made, and what additional evidence can support the decision. CredWise therefore uses the prediction model as one part of a larger decision-support workflow rather than treating the model as the complete system.

\subsection{Explainability and Probability Calibration}
Interpretability is an important issue when machine learning is used for credit-risk assessment. Bussmann et al. \citep{bussmann2021explainable} studied explainable machine learning in credit risk management and discussed interpretability in financial lending using models, visualizations, and summary explanations. Arrieta et al. \citep{arrieta2020xai} provided a broader review of the opportunities and challenges of explainable AI. SHAP \citep{lundberg2017shap} is a widely used method for explaining individual predictions through feature contributions. In CredWise, SHAP is used to identify features that increase or decrease the model output for an applicant and to examine global feature importance.
The reliability of predicted probabilities is another important issue. A classifier may rank applicants well even when its predicted probabilities are not well calibrated \citep{niculescu2005predicting,platt1999probabilistic,guo2017calibration}. The Brier score provides a direct measure of the quality of probabilistic predictions \citep{brier1950verification}. CredWise combines probability calibration with SHAP-based explanations so that the system provides both a calibrated risk probability and information about the features associated with the prediction.

\subsection{Retrieval-Augmented Systems for Decision Support}
A model prediction may not provide all the information needed to make a decision. For example, a reviewer may also need to check a lending policy document. Retrieval-Augmented Generation (RAG) combines information retrieval with language-model based generation \citep{lewis2020rag}. Dense retrieval and sentence embeddings can be used to represent text for retrieval \citep{karpukhin2020dpr,reimers2019sbert}, while FAISS can be used for efficient similarity search \citep{johnson2019faiss}. Different retrieval methods can also be combined using approaches such as Reciprocal Rank Fusion \citep{cormack2009rrf}. This can be useful when policy documents contain both exact terms and semantically related expressions.
Earlier work used FAISS- and BM25-based retrieval together with multiple agents and local language models for enterprise customer support \citep{tiwari2026shopease}. In the present study, retrieval is used for a more specific purpose: retrieving lending-policy evidence and presenting it together with the model prediction and other decision evidence.

\subsection{Agent-Based Systems and Human Oversight}
Language-model based agents have been studied for tasks involving reasoning, tool use, and multi-step interaction. Examples include ReAct \citep{yao2023react}, surveys of autonomous LLM agents \citep{wang2023agentsurvey}, and AgentBench \citep{liu2023agentbench}. In financial decision support, however, additional controls are needed. An agent should not act as an unrestricted decision maker. The system should control which tools the agent can use, what data it can access, and how its output is presented to the human reviewer. This is consistent with guidelines for human-AI interaction \citep{amershi2019guidelines}.
An earlier study on local LLM evaluation found that consistency between automated judgments does not necessarily mean agreement with human ratings \citep{tiwari2026consistency}. This motivates the use of separate and controlled evaluations for the agent outputs in CredWise rather than assuming that consistent agent behaviour is sufficient evidence of reliability.

\subsection{Research Gap}
Existing research provides important foundations for the individual parts of a credit-risk decision system, including prediction \citep{hand1997statistical,lessmann2015benchmarking}, explainability \citep{bussmann2021explainable,lundberg2017shap}, calibration \citep{niculescu2005predicting,guo2017calibration}, retrieval \citep{lewis2020rag}, and agent coordination \citep{yao2023react,wang2023agentsurvey}. However, these components are often studied separately. There is therefore a need to examine how they can be connected in a controlled credit-risk decision-support workflow while keeping the source of each output clear.
CredWise addresses this gap by bringing these components together in one academic research prototype. The language model or agent layer is not used as the final decision maker. Instead, predictions, explanations, retrieved policy evidence, and structured analytics are kept as separate sources of evidence and combined for review by a human decision maker.

\section{Research Questions}
The main aim of this study is to evaluate whether a controlled combination of credit-risk prediction, probability calibration, explainability, policy retrieval, structured analytics, and agent-based workflows can provide useful support to a human reviewer during credit-risk analysis. The evaluation is organized around the following research questions.

\subsection{RQ1: Credit-Risk Prediction Performance}
\textbf{RQ1: How well does the final XGBoost model perform on a temporally separated test period?}
This question evaluates the final model on the 2018 test period using accuracy, precision, recall, F1-score, ROC-AUC, PR-AUC, and Brier score. These metrics cover both classification performance and the quality of the predicted probabilities.

\subsection{RQ2: Feature Set Comparison}
Does adding interest-rate and related loan features improve prediction
performance?
Feature Set A is compared with Feature Set B in a controlled ablation
experiment using the same modelling pipeline. Feature Set B adds
\texttt{int\_rate}, \texttt{installment}, and \texttt{sub\_grade}.
The comparison focuses on changes in ROC-AUC, PR-AUC, precision, recall, F1-score, and Brier score. The final model is then evaluated separately using the temporal train-validation-test setup described in Section~4.

\subsection{RQ3: Probability Calibration}
\textbf{RQ3: Does probability calibration improve the reliability of the model's predicted default probabilities?}
The raw probabilities produced by XGBoost are compared with sigmoid-calibrated probabilities using the Brier score and expected calibration error (ECE). Ranking metrics are also retained to check whether calibration changes the model's ranking behaviour.

\subsection{RQ4: Temporal Stability of Model Explanations}
\textbf{RQ4: Are the feature-importance patterns from SHAP explanations stable across the 2017 and 2018 periods?}
This question examines whether the main features contributing to the model predictions remain similar across the two periods. The analysis uses the Spearman correlation between global feature-importance rankings and the overlap among the top-ranked features.

\subsection{RQ5: Feature Distribution Drift}
\textbf{RQ5: Which input features show the largest distribution changes between the training period and later evaluation periods?}
The Population Stability Index (PSI) is used to compare feature distributions between the 2007--2016 training period and the 2017 and 2018 periods. This identifies features whose distributions changed over time and provides additional context for interpreting temporal model performance.

\subsection{RQ6: Controlled System Component Evaluation}
\textbf{RQ6: Can the policy retrieval, SQL analytics, and agent-routing components produce the expected outputs under controlled evaluation cases?}
The policy retrieval component is evaluated on 28 labeled
queries using Hit@1, Hit@2, Hit@3, Hit@5, and MRR, for BM25, FAISS,
and a reciprocal-rank-fusion hybrid. The SQL analytics component is
evaluated on six benchmark cases covering query generation, execution,
and result matching. The agent-routing component is evaluated on 45
predefined cases covering SQL, policy, risk, and decision-intelligence
paths.

\subsection{RQ7: Integrated Decision-Support Workflow}
\textbf{RQ7: Can the different evidence sources be combined into a single
human-reviewable decision-support output?}
This question examines whether the model prediction, calibrated probability, SHAP evidence, policy evidence, and structured analytics can be combined into a
controlled decision-intelligence output. The evaluation uses one
representative applicant from the 2018 test period and checks the
presence and validity of these components rather than using generated
text itself as a measure of prediction accuracy.
\section{Dataset and Experimental Setup}
\subsection{Dataset}
The experiments use the Lending Club loan dataset, covering loans issued between 2007 and 2018, with information on loan characteristics, borrower characteristics, and credit history.
The target variable is \texttt{loan\_status}: loans marked \textit{Fully Paid} are treated as non-default cases and \textit{Charged Off} loans as default cases; other status categories are excluded from the final binary dataset.
After cleaning and preprocessing, the final dataset contains 1,345,310 loan records and 20 columns, including the target variable, with 18 features used for prediction. Features that could reveal the outcome after loan origination were removed; in particular, \texttt{last\_pymnt\_amnt} was excluded since it reflects repayment activity that occurs after the loan is issued.
The target distribution contains 1,076,751 non-default cases and 268,559 default cases, so the dataset is imbalanced, with about 20\% of observations in the default class.
\begin{figure}[H]
    \centering
    \includegraphics[width=\linewidth]{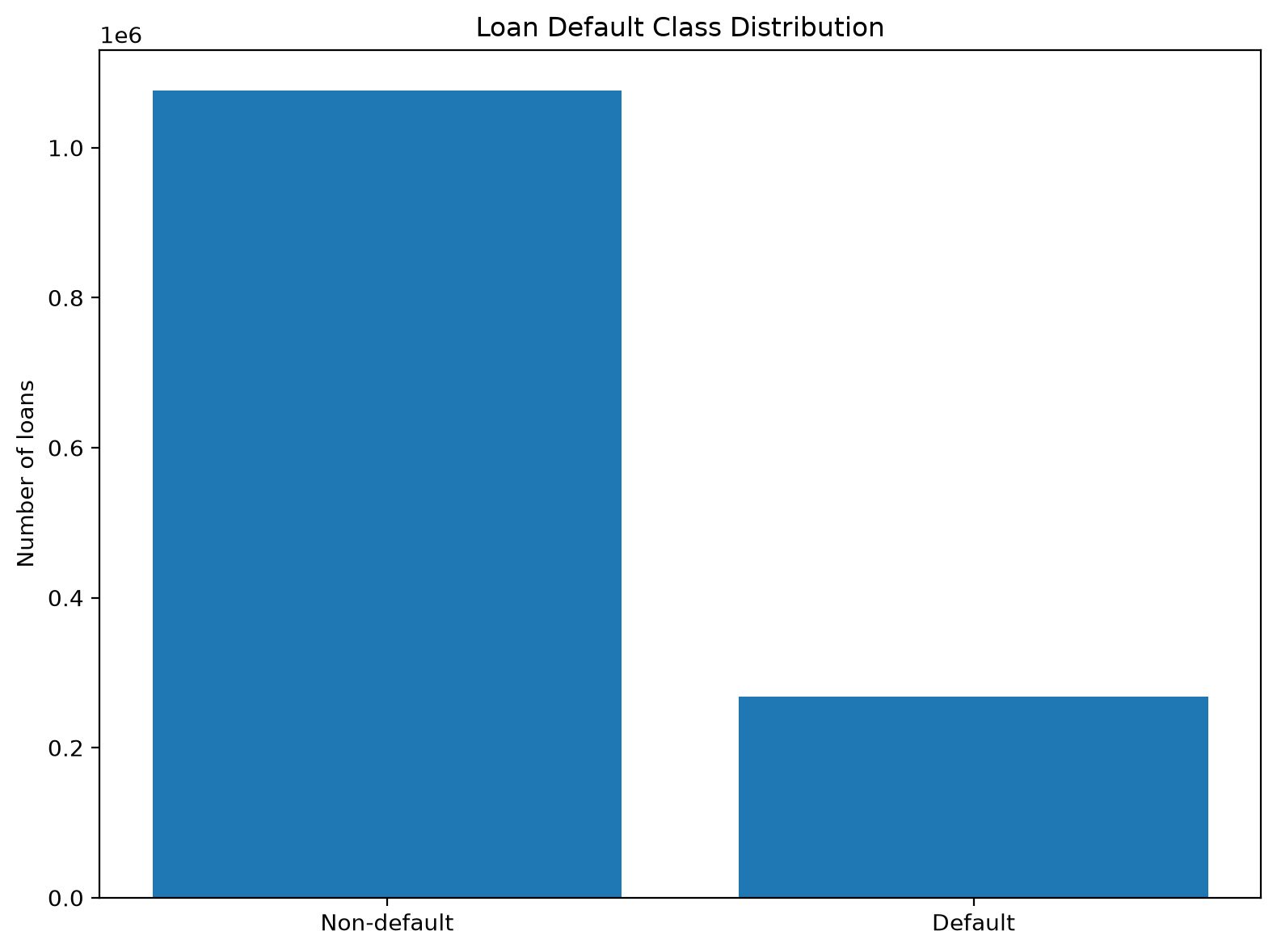}
    \caption{Class distribution in the processed Lending Club dataset.}
    \label{fig:class_distribution}
\end{figure}
Figure~\ref{fig:class_distribution} shows the class distribution; this imbalance was addressed during training using a positive-class weight.
\subsection{Temporal Data Split}
\begin{figure}[H]
    \centering
    \includegraphics[width=\linewidth]{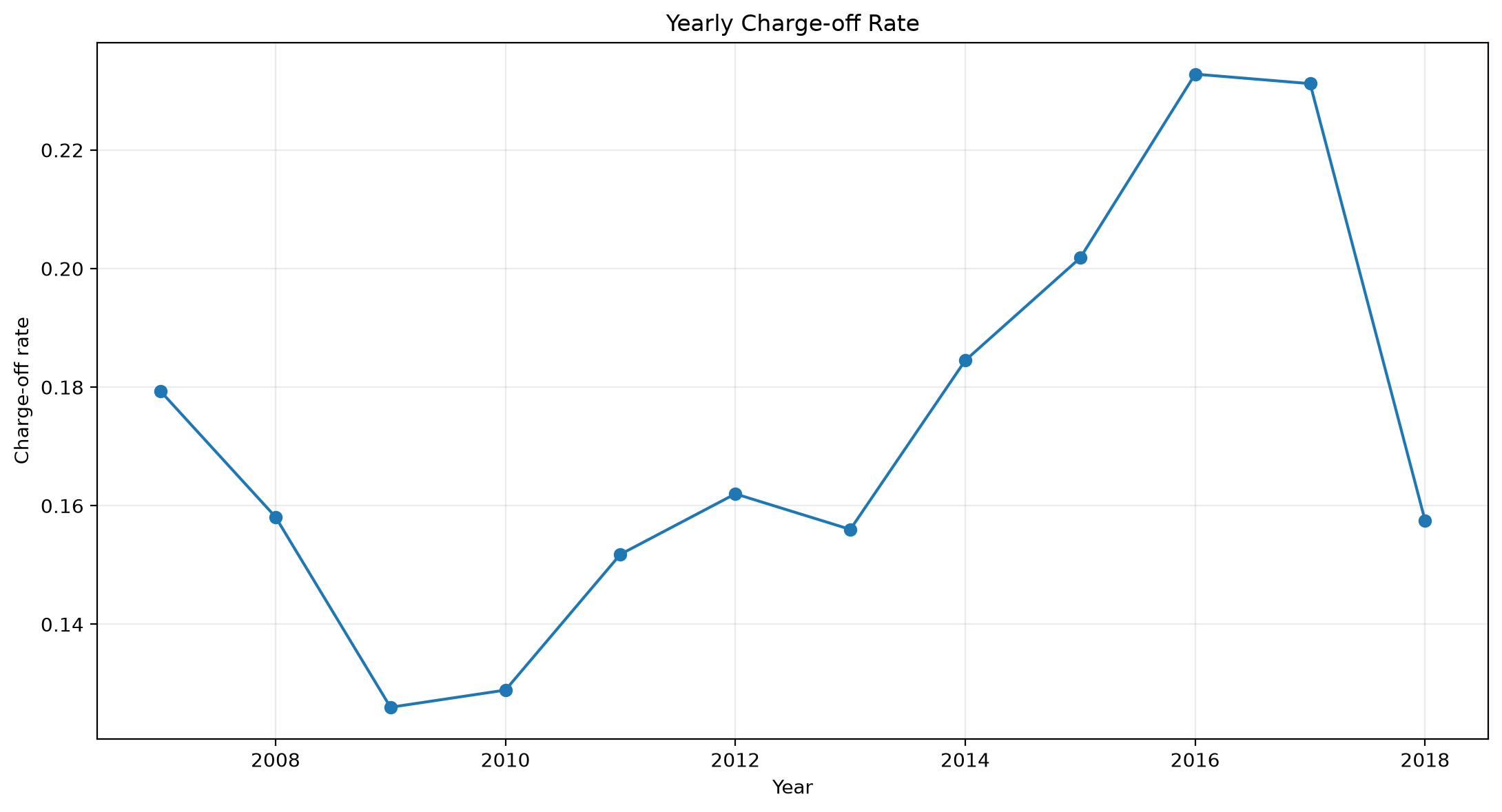}
    \caption{Year-wise charge-off rate in the Lending Club dataset.}
    \label{fig:yearly_chargeoff}
\end{figure}

A temporal split, rather than a random train-test split, was used so the model could be evaluated on a later period than it was trained on: 2007--2016 for training, 2017 for validation, and 2018 as the final test period. The 2018 data was not used for model fitting, hyperparameter selection, or calibration, so the test set represents a genuinely later time period than the training data.
Table~\ref{tab:dataset_split} summarizes the split.
\begin{table}[H]
    \centering
    \caption{Temporal split of the processed dataset.}
    \label{tab:dataset_split}
    \begin{tabular}{lrrr}
        \hline
        \textbf{Period} & \textbf{Purpose} & \textbf{Samples} & \textbf{Default Rate} \\
        \hline
        2007--2016 & Training & 1,119,699 & 19.70\% \\
        2017 & Validation & 169,300 & 23.12\% \\
        2018 & Test & 56,311 & 15.75\% \\
        \hline
        \textbf{Total} & & \textbf{1,345,310} & \\
        \hline
    \end{tabular}
\end{table}
The default rate varies across periods, which is one reason for using a time-based evaluation. Figure~\ref{fig:yearly_chargeoff} shows the yearly charge-off rate.

\subsection{Prediction Features}
Two feature sets were used in the study. Feature Set A contains 15 basic borrower, loan, and credit-history features. Feature Set B adds
\texttt{int\_rate}, \texttt{installment}, and \texttt{sub\_grade}, giving 18 features in total. The two sets are compared in the feature ablation experiment. The final model uses Feature Set B.
\begin{table}[H]
    \centering
    \caption{Features used by the final CredWise model.}
    \label{tab:final_features}
    \begin{tabular}{ll}
        \hline
        \textbf{Feature} & \textbf{Meaning} \\
        \hline
        \texttt{loan\_amnt} & Loan amount \\
        \texttt{term} & Loan term \\
        \texttt{int\_rate} & Interest rate \\
        \texttt{installment} & Monthly payment \\
        \texttt{sub\_grade} & Loan sub-grade \\
        \texttt{emp\_length} & Employment length \\
        \texttt{home\_ownership} & Home ownership \\
        \texttt{annual\_inc} & Annual income \\
        \texttt{verification\_status} & Income verification \\
        \texttt{purpose} & Loan purpose \\
        \texttt{dti} & Debt-to-income ratio \\
        \texttt{delinq\_2yrs} & Recent delinquencies \\
        \texttt{inq\_last\_6mths} & Recent credit inquiries \\
        \texttt{open\_acc} & Open credit accounts \\
        \texttt{pub\_rec} & Public records \\
        \texttt{revol\_bal} & Revolving balance \\
        \texttt{revol\_util} & Credit utilization \\
        \texttt{total\_acc} & Total credit accounts \\
        \hline
    \end{tabular}
\end{table}
The target variable \texttt{loan\_status} is not used as an input feature. Similarly, \texttt{issue\_d} and \texttt{issue\_month} are used only to create the temporal train-validation-test split.

\subsection{Data Preprocessing}
Numerical features were converted to numeric representations and categorical features were encoded, with all preprocessing steps fitted on the training data and then applied to the validation and test periods. The same preprocessing pipeline is stored with the final model so the evaluation transformation can be reproduced at inference time.

\subsection{XGBoost Model}
XGBoost was chosen for its structured numerical and categorical inputs, using gradient-boosted decision trees to produce a default probability. The model was trained on 2007--2016 data, validated on 2017 with early stopping, and used \texttt{scale\_pos\_weight} to account for class imbalance. The best validation iteration was 187, with a validation log-loss of 0.633752.

\subsection{Probability Calibration}
The raw XGBoost probability was calibrated before use in the decision-support layer, using sigmoid calibration for its simplicity and stability as a post-processing step. Calibration was fitted without using the 2018 test outcomes; the calibrated probability is used by the risk engine, while the raw output is retained for comparison. The final decision threshold is 0.23: an applicant is classified as predicted default when the calibrated probability meets or exceeds this value.

\subsection{Evaluation Metrics}
Classification performance is evaluated using accuracy, precision, recall, and F1-score, with ROC-AUC measuring ranking performance and PR-AUC providing an additional view for the imbalanced default class. Probability quality is evaluated using the Brier score and Expected Calibration Error (ECE). For the final 2018 results, bootstrap resampling is used to estimate 95\% confidence intervals, reflecting sampling variability in the test-set metrics.

\subsection{Experimental Reproducibility}
The final model, preprocessing pipeline, probability calibrator, SHAP explainer, and classification threshold are stored as separate artifacts and reloaded during evaluation to verify reproducible outputs. The full evaluation is performed without modifying the 2018 test set, with prediction, calibration, explanation, retrieval, SQL, and agent evaluations treated as independent experiments so each component can be examined separately.

\section{CredWise Framework}
CredWise is an academic decision-support prototype for credit-risk
analysis that combines a machine learning prediction model with
probability calibration, explainability, policy retrieval, SQL-based
analytics, and controlled agent workflows. The main design goal is to
keep these components separate while letting them work together: the
prediction model provides the risk probability, the explainability
component provides model evidence, the policy component provides
relevant policy information, and the SQL component provides structured analytical information, all combined in a final decision-support layer for human review.

\subsection{System Architecture}
\begin{figure}[H]
    \centering
    \includegraphics[width=\linewidth]{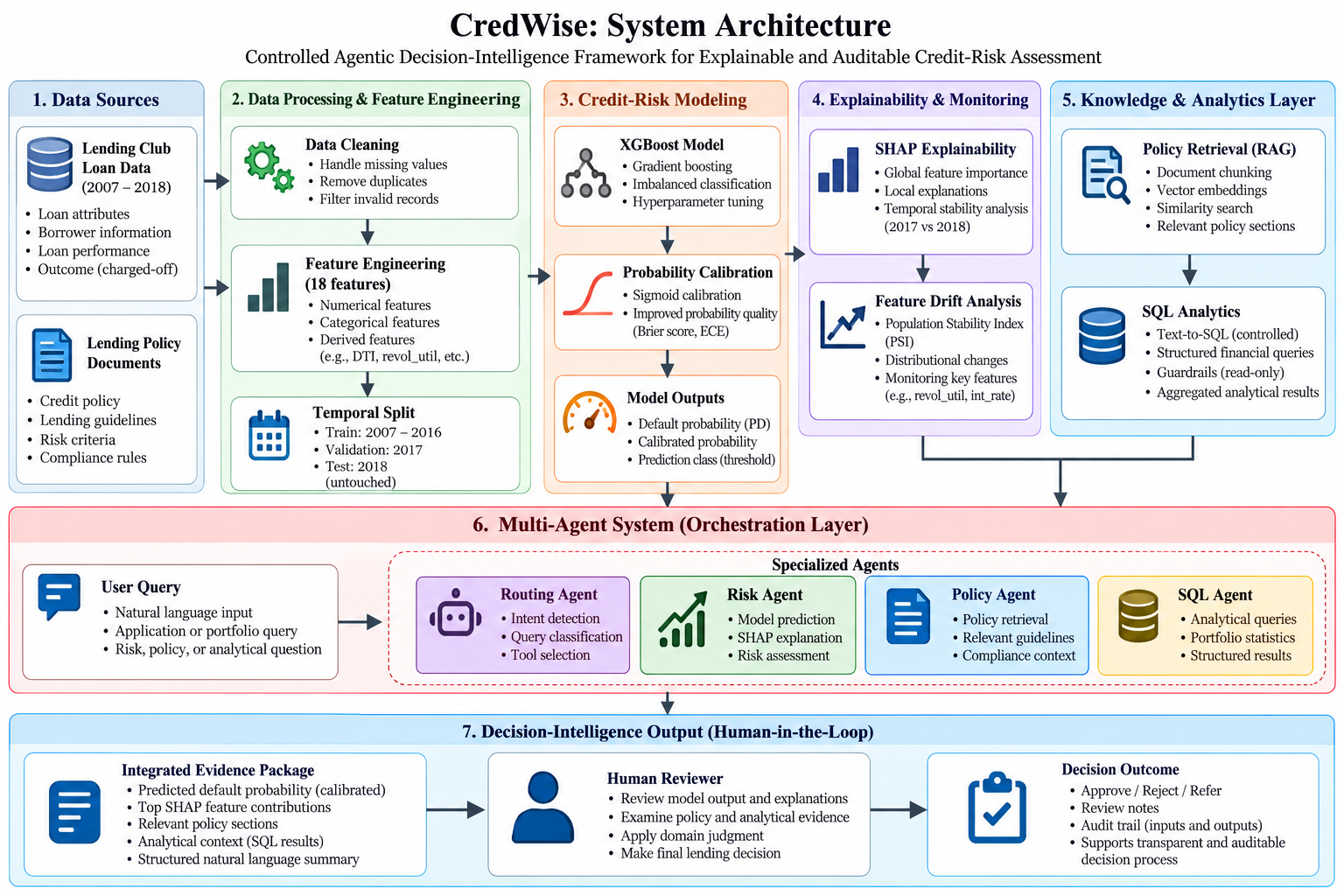}
    \caption{Overall architecture of the CredWise decision-support framework.}
    \label{fig:credwise_architecture}
\end{figure}

The overall architecture, shown in Figure~\ref{fig:credwise_architecture}, has five main stages: input and preprocessing, risk prediction, evidence generation, agent-based coordination, and human review.
Applicant information passes through the same preprocessing pipeline
used during training and is fed to the XGBoost model, which produces a raw default probability. This is calibrated to obtain the probability used by the risk engine, which is then compared against the classification threshold to determine the predicted class and assign a risk level; the prediction itself is not treated as a final lending decision.
In parallel, the SHAP component explains the model prediction, the
policy retrieval component retrieves relevant lending-policy sections, and the SQL analytics component provides structured information from the available data. These outputs are passed to the decision-intelligence layer, whose final output contains the model prediction, model evidence, policy evidence, and analytical context, with the human reviewer remaining responsible for the final decision.

\subsubsection{Research Dashboard}
\begin{figure}[H]
    \centering
    \includegraphics[width=\linewidth]{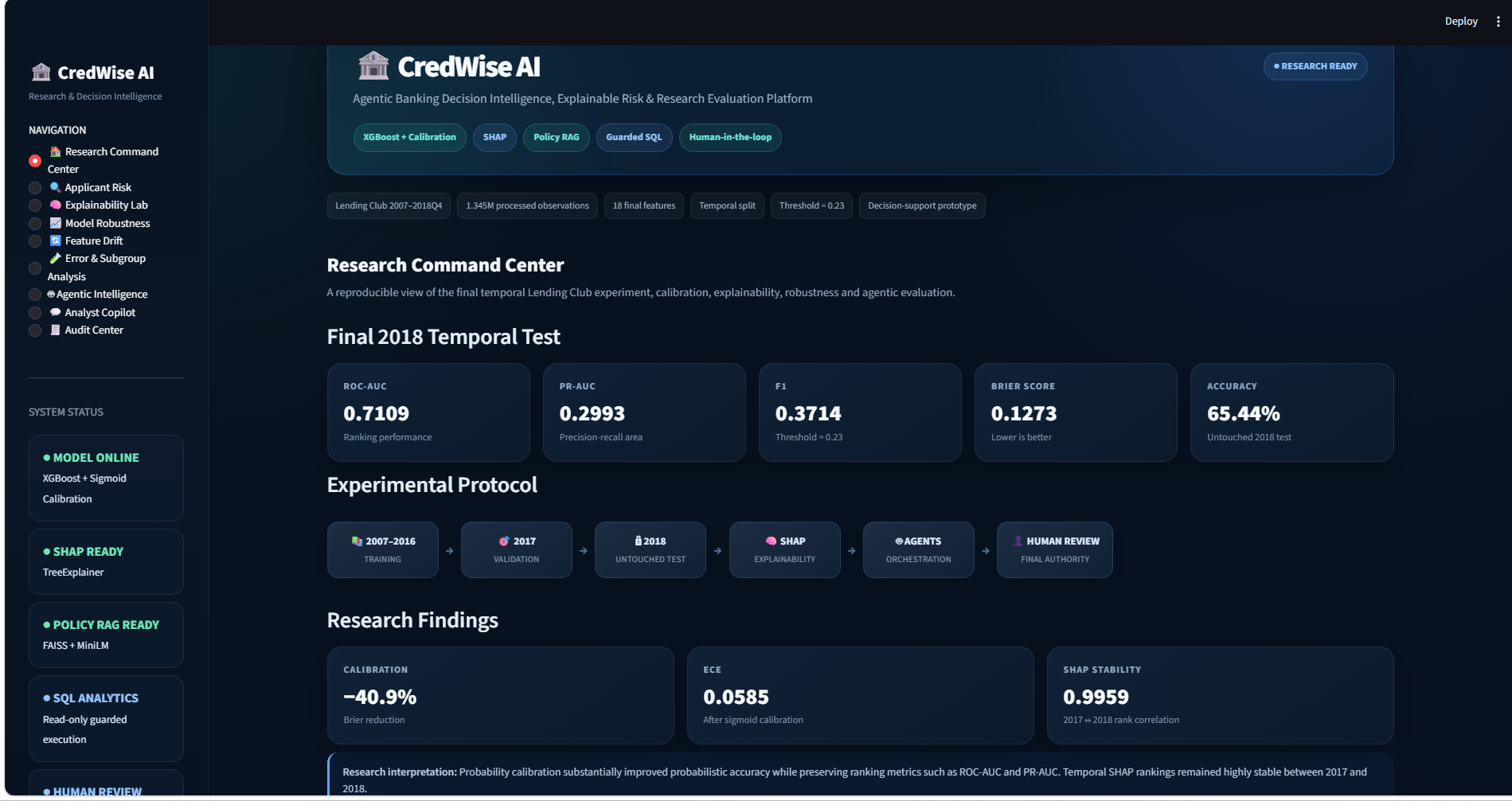}
    \caption{Streamlit-based CredWise research dashboard integrating
    risk prediction, explainability, robustness, feature drift, error
    analysis, agentic intelligence, analytics, and audit information.}
    \label{fig:credwise_dashboard}
\end{figure}

The CredWise prototype provides a Streamlit-based research dashboard
that brings the main outputs of the framework into a single review
interface, as shown in Figure~\ref{fig:credwise_dashboard}. The
dashboard presents the final temporal test results together with
calibration, explainability, model robustness, feature drift, error
and subgroup analysis, agentic intelligence, structured analytics, and audit information.
The dashboard is intended as a research and inspection interface rather than as a replacement for the underlying evaluation procedures. The model prediction, SHAP explanations, policy evidence, analytical results, and evaluation outputs remain separate, while the dashboard allows them to be inspected within a common workflow. It also displays the temporal experimental protocol, including the 2007--2016 training period, 2017 validation period, and 2018 untouched test period, along with the main final-test metrics.
The interface is therefore used to make the experimental outputs easier to inspect and compare during research evaluation. It does not change the underlying model prediction or replace the human review process.

\subsection{Risk Prediction and Calibration}
The risk prediction layer uses the final XGBoost model described in
Section~4. It receives the 18 selected features $\mathbf{x}$ and
produces a raw probability:
\begin{equation}
    p_{\mathrm{raw}} = f_{\mathrm{XGB}}(\mathbf{x}),
\end{equation}
where $f_{\mathrm{XGB}}$ is the trained model. The raw probability is
not used directly and is instead passed to the calibration layer.
The calibration layer converts the raw probability into a calibrated
probability using a fitted sigmoid function $g(\cdot)$:
\begin{equation}
    p_{\mathrm{cal}} = g(p_{\mathrm{raw}}).
\end{equation}
Calibration is evaluated separately from the classification model so
that changes in probability quality can be measured without affecting
the underlying ranking. With a classification threshold of 0.23,
\begin{equation}
\hat{y} =
\begin{cases}
1, & p_{\mathrm{cal}} \geq 0.23,\\
0, & p_{\mathrm{cal}} < 0.23,
\end{cases}
\end{equation}
where $\hat{y}=1$ denotes predicted default and $\hat{y}=0$ predicted
non-default.
The system also assigns a risk level from the calibrated probability:
\begin{equation}
\text{Risk Level} =
\begin{cases}
\text{Low Risk}, & p_{\mathrm{cal}} < 0.15,\\
\text{Medium Risk}, & 0.15 \leq p_{\mathrm{cal}} < 0.30,\\
\text{High Risk}, & p_{\mathrm{cal}} \geq 0.30.
\end{cases}
\end{equation}
These ranges organize the decision-support output only and are not
presented as regulatory or industry-standard categories.
\subsection{Explainability and Evidence Generation}
The explainability layer uses SHAP to describe each feature's
contribution to a model output \citep{lundberg2017shap}. For an
applicant,
\begin{equation}
    f(\mathbf{x}) = \phi_0 + \sum_{i=1}^{M}\phi_i,
\end{equation}
where $\phi_0$ is the base value and $\phi_i$ is feature $i$'s
contribution. These values identify features that raise or lower the
model output and, when aggregated across the test set, support global
analysis. Explanations are treated as model evidence, not as
independent causal claims about borrower behaviour.
The explanation component therefore provides both individual-level
evidence and global feature-importance information. The individual
contributions are used to identify risk-raising and risk-reducing
features for a given applicant, while aggregated SHAP values are used
to study feature importance and temporal stability.

\subsection{Policy Retrieval and SQL Analytics}
The policy document is divided into sections, each represented as a
vector using a sentence embedding model and indexed in FAISS for
similarity search \citep{reimers2019sbert,johnson2019faiss}. For a
user query, the retrieval component searches the index and returns
relevant sections with metadata, such as section and source,
preserving a link between retrieved evidence and its source. This
layer is intentionally separate from the prediction model, so a
retrieved policy statement does not change the XGBoost probability but
instead provides additional evidence for the reviewer.
The SQL analytics layer supports structured analytical questions, such
as portfolio-level counts or average predicted risk, through
controlled, read-only query execution. Only approved tables and
operations are allowed, while destructive operations, unauthorized
tables, and multi-statement queries are blocked, reducing the risk
that a language model or user query can modify the underlying database. SQL output is treated as analytical evidence and is kept separate from the model prediction.
Together, the policy retrieval and SQL layers provide two additional
sources of evidence: policy information from the indexed lending-policy document and structured analytical information from the available data. Neither component changes the underlying model prediction.

\subsection{Controlled Agent Workflow and Decision Intelligence}
A supervisor/routing layer identifies the type of each request and
directs it to the appropriate component---for example, a policy
question to the policy component, a portfolio-level numerical question
to the SQL component, or a ``why'' question to the risk and
explainability components:
\begin{equation}
    \text{User Query}
    \rightarrow
    \text{Supervisor/Router}
    \rightarrow
    \text{Specialized Component}.
\end{equation}
Using specialized components rather than one general-purpose agent
makes the source of each part of the final response easier to
identify. This workflow draws on prior work on reasoning and tool-use
agents \citep{yao2023react,wang2023agentsurvey}, but CredWise uses a
controlled version suited to a financial decision-support setting.
The decision-intelligence layer assembles a structured package for
each applicant, containing:
\begin{itemize}
    \item calibrated probability of default and predicted class;
    \item assigned risk level and model/prediction information;
    \item SHAP-based risk-raising and risk-reducing features;
    \item retrieved policy evidence and relevant SQL-based analytics; and
    \item a structured summary for human review.
\end{itemize}
This layer organizes evidence rather than producing an independent
prediction, preserving the distinction between model, policy, and
analytical evidence. The final output therefore connects the
specialized components without treating the agent or language model as
the source of the underlying model prediction.

\subsection{Human-in-the-Loop and Auditability}
CredWise supports review but does not make the final lending decision,
following the principle that AI systems in important workflows should
keep users informed and support human control \citep{amershi2019guidelines}, consistent with the emphasis on interpretable, understandable outputs in financial lending \citep{bussmann2021explainable,bucker2022lending}.
The reviewer considers the model probability, explanation, policy
evidence, and analytical context together; no single evidence source
is presented as sufficient on its own.
CredWise keeps four types of information distinct:
\begin{enumerate}
    \item \textbf{Model evidence:} prediction probability and SHAP
    feature contributions.
    \item \textbf{Policy evidence:} retrieved sections from the
    lending policy document.
    \item \textbf{Analytical evidence:} results from the controlled
    SQL analytics layer.
    \item \textbf{Decision-support output:} a structured summary
    combining the available evidence for human review.
\end{enumerate}
This separation prevents a generated explanation from being mistaken
for a model output or policy statement and makes each component easier
to evaluate independently. Each component is evaluated separately before combination: the prediction model on temporal test data, calibration using probability-based metrics, SHAP for feature importance and temporal stability, policy retrieval using labeled queries, SQL using predefined analytical cases, and the agent layer using predefined routing cases. The decision-intelligence layer is then checked to confirm that the expected evidence fields are present and correctly connected. This component-wise strategy is important because a successful final response alone does not confirm that every underlying component is working correctly; detailed results are presented in Section~6.
\section{Experimental Results}
This section reports the experimental results of the CredWise framework. The evaluation is divided into model performance, feature ablation, probability calibration, explanation stability, feature drift, error analysis, subgroup analysis, and evaluation of the retrieval, SQL, and agent components.
The final model is evaluated on the 2018 test period, which was not used for model training or model selection. Unless otherwise stated, the reported test results therefore refer to this 2018 period.

\subsection{Overall Prediction Performance}
\begin{figure}[H]
    \centering
    \includegraphics[width=\linewidth]{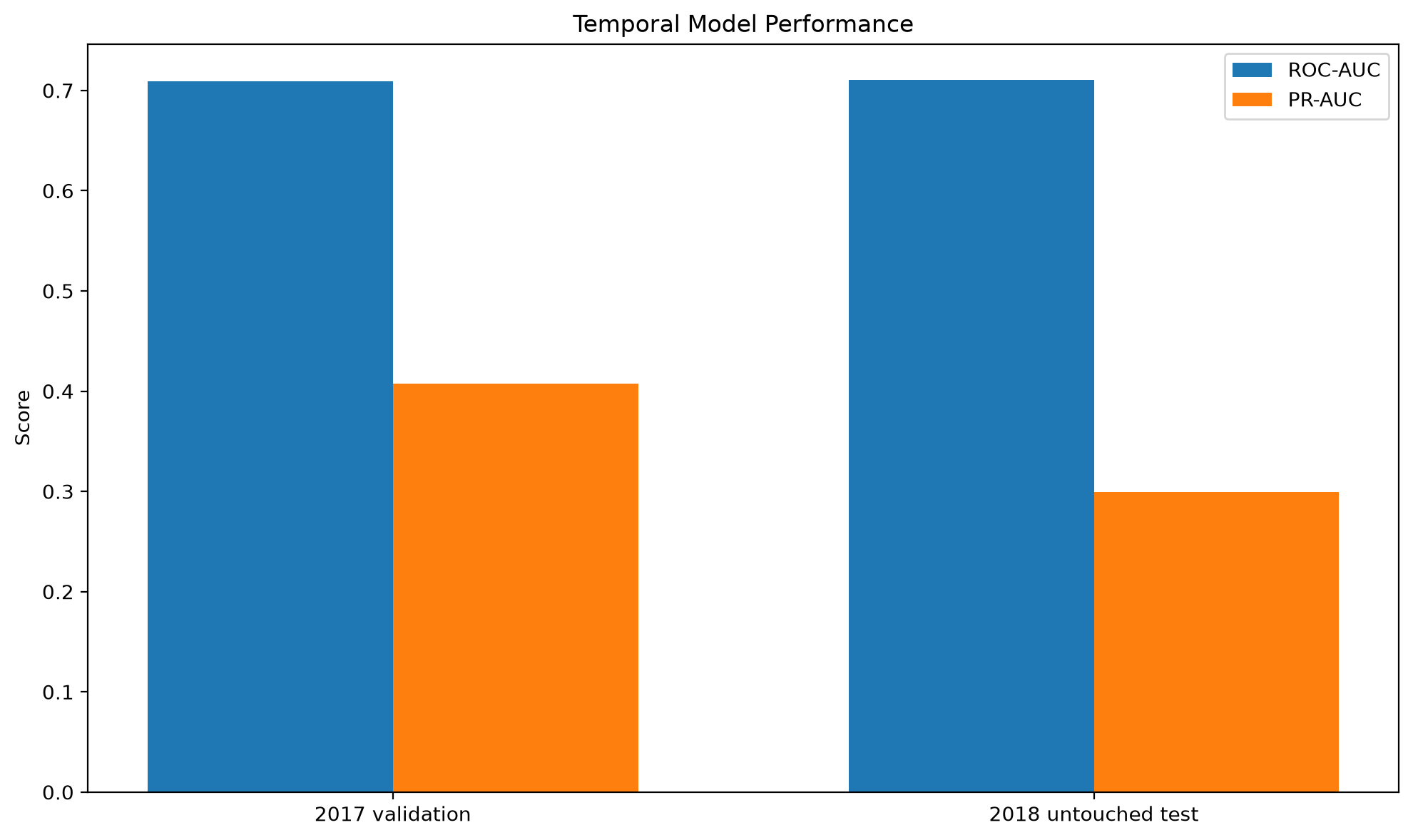}
    \caption{Comparison of model performance between the 2017 validation period and the 2018 test period.}
    \label{fig:temporal_performance}
\end{figure}

The final XGBoost model was trained using data from 2007--2016 and
validated on the 2017 period. Early stopping selected iteration 187,
where the validation log-loss was 0.633752. On the 2018 test set, the calibrated model obtained an accuracy of 65.44\%, precision of 26.02\%, recall of 64.84\%, and F1-score of 0.3714. The ROC-AUC was 0.7109 and the PR-AUC was 0.2993. The Brier score after calibration was 0.1273.
The results for the 2017 validation period and the 2018 test period are shown in Table~\ref{tab:temporal_performance}.
\begin{table}[H]
    \centering
    \caption{Performance of the final model on the 2017 validation and 2018 test periods.}
    \label{tab:temporal_performance}
    \begin{tabular}{lrr}
        \hline
        \textbf{Metric} & \textbf{2017} & \textbf{2018} \\
        \hline
        Accuracy & 0.6403 & 0.6544 \\
        Precision & 0.3539 & 0.2602 \\
        Recall & 0.6724 & 0.6484 \\
        F1-score & 0.4637 & 0.3714 \\
        ROC-AUC & 0.7091 & 0.7109 \\
        PR-AUC & 0.4075 & 0.2993 \\
        Brier score & 0.1606 & 0.1273 \\
        \hline
    \end{tabular}
\end{table}
Figure~\ref{fig:temporal_performance} provides a visual comparison of
the main performance metrics across the two periods.
The ROC-AUC values are similar across the two periods, while the PR-AUC and F1-score are lower in 2018. This difference is partly associated with the different class distribution in the two periods. Therefore, performance is reported using several metrics rather than a single measure.

\subsection{Feature Set Ablation}
Two feature sets were evaluated to measure the effect of adding
interest-rate and related loan features. Feature Set A contains 15
features, while Feature Set B adds \texttt{int\_rate},
\texttt{installment}, and \texttt{sub\_grade}. This ablation uses a random train-test split (rather than the temporal split used elsewhere in this paper) to isolate the effect of the added features without confounding it with the
training/test-period difference. 
Table~\ref{tab:feature_ablation} presents the results.

\begin{table}[H]
    \centering
    \caption{Feature set ablation results on the earlier random
    train-validation-test split.}
    \label{tab:feature_ablation}
    \begin{tabular}{lrrr}
        \hline
        \textbf{Metric} & \textbf{Feature Set A} & \textbf{Feature Set B} & \textbf{Change} \\
        \hline
        Accuracy & 0.6462 & 0.6497 & +0.0035 \\
        Precision & 0.3113 & 0.3215 & +0.0102 \\
        Recall & 0.6370 & 0.6799 & +0.0429 \\
        F1-score & 0.4182 & 0.4366 & +0.0184 \\
        ROC-AUC & 0.6985 & 0.7198 & +0.0213 \\
        PR-AUC & 0.3650 & 0.3842 & +0.0192 \\
        Brier score & 0.2195 & 0.2140 & -0.0054 \\
        \hline
    \end{tabular}
\end{table}
Feature Set B gives higher ROC-AUC, PR-AUC, recall, precision, and F1-score than Feature Set A. Its Brier score is also lower by 0.0054. The results indicate that the three added features provide useful predictive information in this ablation experiment.
The ablation experiment and the final model evaluation use different
evaluation setups. The ablation results are obtained from the earlier random split, whereas the final model is evaluated using the temporal split with 2007--2016 for training, 2017 for validation, and 2018 for testing. Therefore, the Feature Set B ROC-AUC of 0.7198 in the ablation experiment should not be directly compared with the final temporal-test ROC-AUC of 0.7109. The difference does not result from probability calibration, since calibration does not change the ranking of model scores.
This experiment is an ablation comparison within the specified modelling setup. It does not establish that the added features have a causal effect on loan outcomes.

\subsection{Probability Calibration}
The raw XGBoost probability was compared with the calibrated probability on the 2018 test period. Before calibration, the Brier score was 0.2157 and the ECE was 0.2862. After sigmoid calibration, the Brier score decreased to 0.1273 and the ECE decreased to 0.0585.
Table~\ref{tab:calibration_results} summarizes the calibration results.
\begin{table}[H]
    \centering
    \caption{Probability calibration results on the 2018 test period.}
    \label{tab:calibration_results}
    \begin{tabular}{lrr}
        \hline
        \textbf{Metric} & \textbf{Before Calibration} & \textbf{After Calibration} \\
        \hline
        Brier score & 0.2157 & 0.1273 \\
        ECE & 0.2862 & 0.0585 \\
        \hline
    \end{tabular}
\end{table}
The Brier score decreased by 0.0883 and the ECE decreased by 0.2278.
The calibration step does not change the ordering of model scores.
Therefore, the calibration result is interpreted as an improvement in
probability quality rather than as a change in the underlying ranking
model.
For additional context, the default rate in the 2018 test set was 0.1575. A constant prediction equal to this test-set prevalence gives a Brier score of approximately 0.1327. The calibrated model therefore has a Brier score of 0.1273, corresponding to a Brier skill score of approximately 0.040 relative to this prevalence baseline. This comparison shows that the calibration result should be interpreted together with a simple prevalence baseline rather than only as a reduction from the raw XGBoost score.

\subsection{Bootstrap Confidence Intervals}
Bootstrap resampling was used to estimate 95\% confidence intervals for the final 2018 test metrics. The results are reported in
Table~\ref{tab:bootstrap_ci}.
\begin{table}[H]
    \centering
    \caption{Bootstrap estimates and 95\% confidence intervals for the 2018 test results.}
    \label{tab:bootstrap_ci}
    \begin{tabular}{lrr}
        \hline
        \textbf{Metric} & \textbf{Estimate} & \textbf{95\% CI} \\
        \hline
        Accuracy & 0.6544 & [0.6505, 0.6584] \\
        Precision & 0.2602 & [0.2544, 0.2661] \\
        Recall & 0.6484 & [0.6380, 0.6579] \\
        F1-score & 0.3714 & [0.3644, 0.3783] \\
        ROC-AUC & 0.7109 & [0.7050, 0.7164] \\
        PR-AUC & 0.2993 & [0.2904, 0.3080] \\
        Brier score & 0.1273 & [0.1258, 0.1289] \\
        \hline
    \end{tabular}
\end{table}

The intervals provide an estimate of the variation of the reported
metrics under bootstrap resampling of the 2018 test observations. They are not intended to represent variation across different datasets or future economic conditions.

\subsection{SHAP Feature Importance and Temporal Stability}
\begin{figure}[H]
    \centering
    \includegraphics[width=\linewidth]{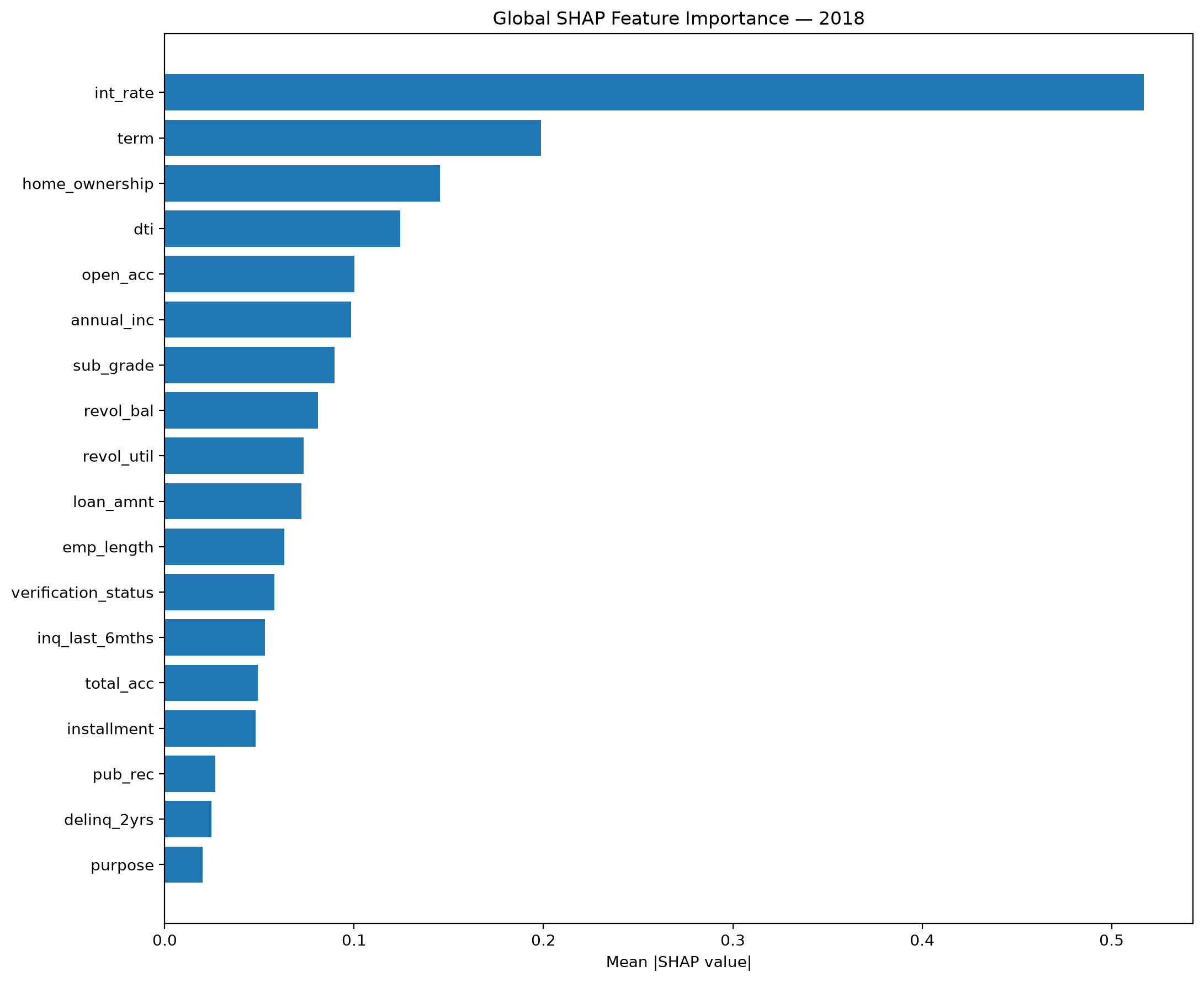}
    \caption{Global SHAP feature importance for the final model on the 2018 test period.}
    \label{fig:shap_importance}
\end{figure}

SHAP was used to examine the contribution of individual features to the model predictions. The global feature-importance analysis for 2018 is shown in Figure~\ref{fig:shap_importance}.
The largest mean absolute SHAP values in 2018 were observed for
\texttt{int\_rate}, \texttt{term}, \texttt{home\_ownership},
\texttt{dti}, and \texttt{open\_acc}. Other features with notable
contributions included \texttt{annual\_inc}, \texttt{sub\_grade},
\texttt{revol\_bal}, \texttt{revol\_util}, and \texttt{loan\_amnt}.
The temporal stability of these rankings was evaluated by comparing the 2017 and 2018 global SHAP rankings. Table~\ref{tab:shap_stability} summarizes the results.
\begin{table}[H]
    \centering
    \caption{Temporal stability of global SHAP feature rankings.}
    \label{tab:shap_stability}
    \begin{tabular}{lr}
        \hline
        \textbf{Measure} & \textbf{Result} \\
        \hline
        Spearman rank correlation & 0.9959 \\
        Spearman $p$-value & $4.18 \times 10^{-18}$ \\
        Top-5 overlap & 100\% \\
        Top-10 overlap & 90\% \\
        Top-15 overlap & 100\% \\
        \hline
    \end{tabular}
\end{table}
The Spearman correlation of 0.9959 indicates that the global feature
ranking was very similar between the two periods. The top-5 and top-15 sets were identical, while the top-10 sets had 90\% overlap. This
result describes stability of the model's feature-importance ranking;
it does not imply that the relationships between the features and
default are causal.

\subsection{Feature Drift Analysis}
\begin{figure}[H]
    \centering
    \includegraphics[width=\linewidth]{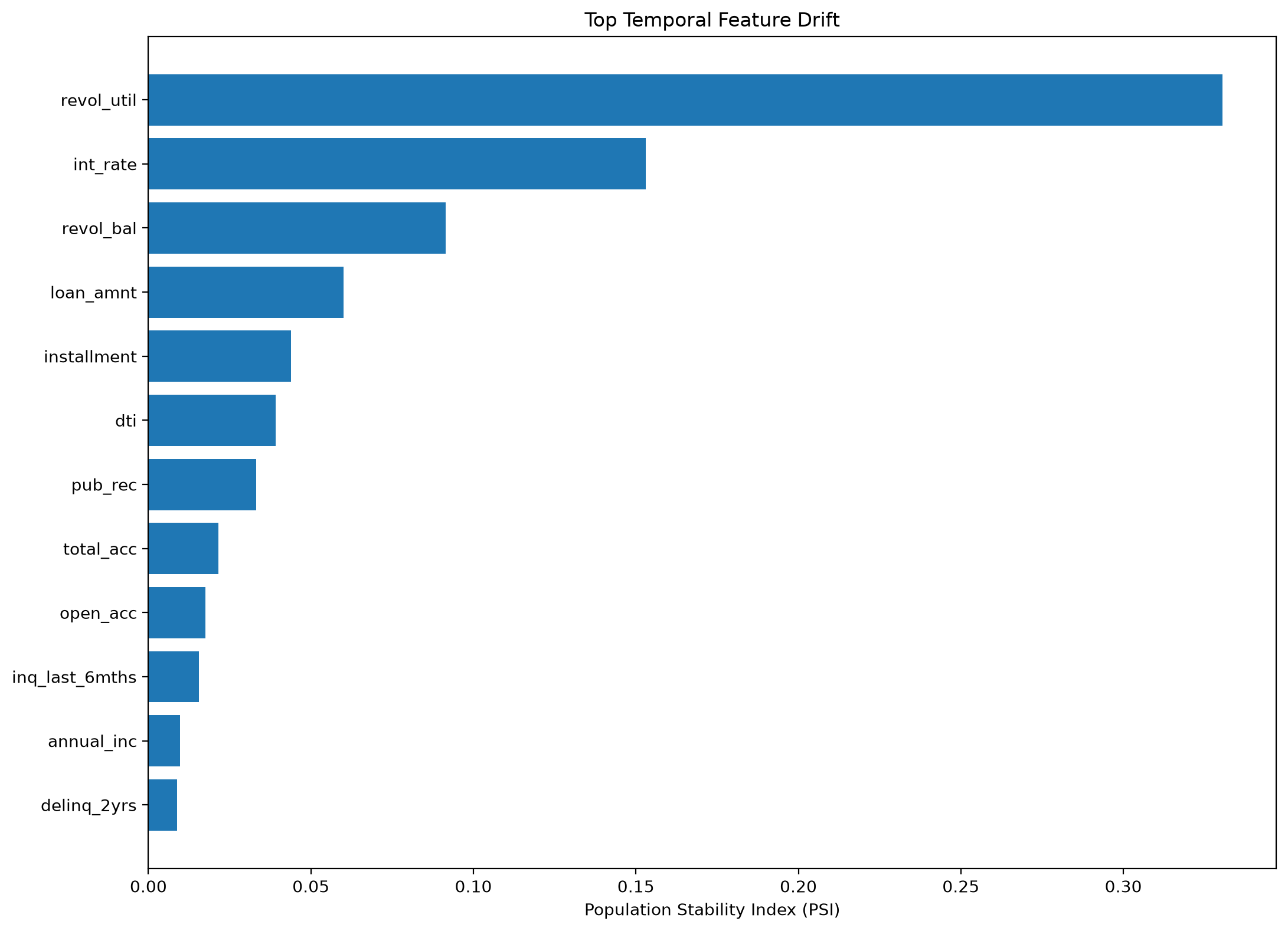}
    \caption{Population Stability Index for selected features between the training period and later evaluation periods.}
    \label{fig:feature_drift}
\end{figure}

Feature distribution changes were measured using the Population
Stability Index (PSI). The 2007--2016 training period was compared with the 2017 and 2018 periods.
Table~\ref{tab:feature_drift} presents selected features with their PSI values.
\begin{table}[H]
    \centering
    \caption{Selected feature drift measurements using PSI.}
    \label{tab:feature_drift}
    \begin{tabular}{lrr}
        \hline
        \textbf{Feature} & \textbf{2017 PSI} & \textbf{2018 PSI} \\
        \hline
        \texttt{revol\_util} & 0.0967 & 0.3305 \\
        \texttt{int\_rate} & 0.0741 & 0.1531 \\
        \texttt{revol\_bal} & 0.0200 & 0.0915 \\
        \texttt{loan\_amnt} & 0.0267 & 0.0600 \\
        \texttt{installment} & 0.0259 & 0.0439 \\
        \texttt{dti} & 0.0024 & 0.0392 \\
        \texttt{pub\_rec} & 0.0000 & 0.0332 \\
        \texttt{total\_acc} & 0.0099 & 0.0216 \\
        \texttt{open\_acc} & 0.0038 & 0.0175 \\
        \texttt{inq\_last\_6mths} & 0.0091 & 0.0155 \\
        \texttt{annual\_inc} & 0.0057 & 0.0097 \\
        \texttt{delinq\_2yrs} & 0.0000 & 0.0089 \\
        \hline
    \end{tabular}
\end{table}
Figure~\ref{fig:feature_drift} shows the PSI values for the selected
features.
The largest shift was observed for \texttt{revol\_util}, whose PSI
increased from 0.0967 in 2017 to 0.3305 in 2018. Interest rate also
showed a noticeable increase, from 0.0741 to 0.1531. These values
indicate that the distributions of some input variables changed over
time.
The drift analysis does not by itself establish model degradation or
the cause of the observed distribution changes. It is therefore
considered together with the temporal performance and error analyses.

\subsection{Error Analysis}
At the selected classification threshold of 0.23, the 2018 test set
produced the confusion matrix shown in
Table~\ref{tab:error_analysis}.
\begin{table}[H]
    \centering
    \caption{Confusion matrix and error rates on the 2018 test period.}
    \label{tab:error_analysis}
    \begin{tabular}{lrr}
        \hline
        & \textbf{Predicted Non-default} & \textbf{Predicted Default} \\
        \hline
        \textbf{Actual Non-default} & 31,102 & 16,342 \\
        \textbf{Actual Default} & 3,118 & 5,749 \\
        \hline
    \end{tabular}
    \vspace{2mm}
    \begin{tabular}{lr}
        \hline
        \textbf{Measure} & \textbf{Value} \\
        \hline
        Non-default error rate & 34.44\% \\
        Default error rate & 35.16\% \\
        \hline
    \end{tabular}
\end{table}
The model produced 16,342 false positives and 3,118 false negatives.
The false-positive group contained, on average, larger loan amounts,
higher interest rates, higher DTI, and higher revolving utilization
than the true-negative group. The false-negative group generally had
lower interest rates and lower DTI than the true-positive group.
These patterns show that some observations are difficult to classify
using the available features. The error analysis is descriptive and
does not imply that any individual feature causes the prediction error.

\subsection{Subgroup Robustness}
Subgroup analysis was performed for employment length, home ownership, loan term, and verification status. The purpose was to examine whether model performance was reasonably consistent across different groups. The subgroup analysis uses the same 2018 test predictions and the same classification threshold used for the main evaluation. The complete subgroup results are reported in the corresponding evaluation table included with the research artifacts.
The subgroup results are descriptive and should not be interpreted as a formal fairness assessment. The analysis is limited to the available group definitions and the observed test-set sample sizes.

\subsection{Policy Retrieval Evaluation}
The policy retrieval component was evaluated using a labeled set of
28 queries covering all nine policy sections. Each query was associated with an expected policy section. The evaluation measured whether the expected section appeared in the top-$k$ retrieved results, for BM25, FAISS, and a reciprocal-rank-fusion hybrid of the two. The agent-routing component was evaluated on 45 labeled requests covering four route types (decision, policy, risk, and SQL). Table~\ref{tab:component_evaluation} reports the retrieval metrics
together with the SQL and agent-routing evaluations.

\begin{table}[H]
    \centering
    \caption{Evaluation of policy retrieval, SQL analytics, and agent routing.}
    \label{tab:component_evaluation}
    \begin{tabular}{lll}
        \hline
        \textbf{Component} & \textbf{Metric} & \textbf{Result} \\
        \hline
        Policy RAG (BM25)   & Hit@1 / Hit@2 / Hit@3 / Hit@5 & 0.714 / 0.857 / 0.857 / 1.000 \\
        Policy RAG (BM25)   & MRR & 0.818 \\
        Policy RAG (FAISS)  & Hit@1 / Hit@2 / Hit@3 / Hit@5 & 0.929 / 1.000 / 1.000 / 1.000 \\
        Policy RAG (FAISS)  & MRR & 0.964 \\
        Policy RAG (Hybrid) & Hit@1 / Hit@2 / Hit@3 / Hit@5 & 0.821 / 1.000 / 1.000 / 1.000 \\
        Policy RAG (Hybrid) & MRR & 0.911 \\
        SQL Agent & SQL exact match & 1.000 \\
        SQL Agent & Execution success & 1.000 \\
        SQL Agent & Result match & 1.000 \\
        Agent Routing & Correct cases & 43/45 (0.956) \\
        \hline
    \end{tabular}
\end{table}
The retrieval component was evaluated on a 28-query labeled set covering all nine policy sections. FAISS achieved the strongest ranking quality (Hit@1 = 0.929, MRR = 0.964), reaching Hit@2 = 1.0. BM25 reached Hit@1 = 0.714 and did not reach Hit@3 = 1.0, needing the full top-5 to retrieve the expected section for every query (MRR = 0.818). The hybrid reciprocal-rank-fusion method fell between the two on Hit@1 (0.821) and MRR (0.911) while matching FAISS from Hit@2 onward. All three methods reached Hit@5 = 1.0. The SQL benchmark contained six cases and achieved exact match, execution success, and result match values of 1.0. The agent-routing evaluation contained 45 predefined cases covering decision, policy, risk, and SQL routes and achieved an overall accuracy of 0.956 (macro F1 = 0.96). Per-route performance was strongest for policy routing (precision = recall = 1.00) and weakest for the risk route, where precision was 0.91 because some SQL-route cases were misclassified as risk; recall for SQL routing was correspondingly 0.90.
These results should be interpreted within the size and design of the
evaluation sets. The 28-query policy benchmark and the 45-case
routing benchmark are larger than a handful of cases but remain
controlled evaluations drawn from a single policy document and a
predefined case set, and should not be treated as evidence of
large-scale retrieval or routing performance in production settings.

\subsection{Decision-Intelligence Evaluation}
The final decision-intelligence component was evaluated using one
representative applicant from the 2018 test period. The applicant had a raw model probability of 0.9004 and a calibrated probability of 0.6108. Since the calibrated probability was above the classification threshold of 0.23, the predicted class was default. The corresponding risk level was High Risk according to the predefined risk ranges.
The final decision-support package contained the calibrated prediction, raw prediction, classification threshold, model information, four risk-raising SHAP features, three risk-reducing SHAP features, two policy-evidence items, and four analytical context fields. The validation checks for the final decision-intelligence output passed.
This evaluation verifies that the required evidence components can be
assembled for a representative case. It does not measure the reliability of the decision-intelligence workflow across a larger population of applicants.

\section{Discussion}
The experiments show that CredWise can combine credit-risk prediction, probability calibration, explainability, policy retrieval, SQL analytics, and controlled agent workflows into a single decision-support system. The results also show that no single metric is sufficient to describe the system. Prediction performance, probability quality, explanation stability, feature drift, and component-level behaviour provide different information.

\subsection{Predictive Performance and Calibration}
The final XGBoost model achieved a ROC-AUC of 0.7109, PR-AUC of 0.2993, and F1-score of 0.3714 on the 2018 test period at the selected threshold of 0.23. These results indicate useful ranking performance, but also show that the model does not perfectly separate default and non-default cases. This is expected in credit-risk prediction, where many factors affecting borrower outcomes may not be available in the dataset.
The PR-AUC result is particularly relevant because the positive class is imbalanced. Reporting both ROC-AUC and precision-recall measures provides a more complete view of classification performance
\citep{davis2006precision,lessmann2015benchmarking}.
The feature ablation experiment showed that Feature Set B performed
better than Feature Set A in the tested setup. ROC-AUC increased from
0.6985 to 0.7198, recall from 0.6370 to 0.6799, and F1-score from 0.4182 to 0.4366.
(This comparison uses a random split and is therefore not directly
comparable in absolute terms to the temporal-split results in Table~3.)
The additional features were \texttt{int\_rate},
\texttt{installment}, and \texttt{sub\_grade}. This suggests that these variables contain useful predictive information in this dataset, but the result does not imply that they cause default outcomes. 
Probability calibration reduced the Brier score from 0.2157 to 0.1273
and ECE from 0.2862 to 0.0585. Since calibration is applied after model prediction, it does not change the underlying XGBoost model or its feature ranking. The result therefore indicates improved probability quality rather than improved underlying classification performance. For context, the Brier score of a constant predictor based on the 2018 test-set default rate is approximately 0.1327, giving a Brier skill score of approximately 0.040 for the calibrated model. Thus, the calibrated probabilities are only modestly better than the prevalence baseline, even though the improvement
over the raw XGBoost probabilities is substantial.
\citep{niculescu2005predicting,platt1999probabilistic,guo2017calibration}. Using both discrimination and calibration measures is useful when model probabilities are used as part of decision support.

\subsection{Temporal Stability and Feature Drift}
The global SHAP rankings for 2017 and 2018 had a Spearman correlation of 0.9959. The top-5 and top-15 feature sets had 100\% overlap, while the top-10 sets had 90\% overlap. This indicates that the relative importance of the main features was highly similar across the two evaluated periods. However, this result does not establish stability under future conditions. SHAP values describe model feature contributions and should not be interpreted as causal effects \citep{lundberg2017shap,bussmann2021explainable}.
The feature drift analysis showed that the input distributions changed between the training period and later periods. The largest observed PSI was for \texttt{revol\_util}, reaching 0.3305 in 2018, followed by \texttt{int\_rate} with a PSI of 0.1531. These results indicate temporal changes in the input data, but PSI alone does not establish the reasons for those changes or prove model degradation. The combination of temporal performance, SHAP stability, and feature drift therefore provides a broader view of model behaviour
\citep{gama2014conceptdrift}.

\subsection{Error and Subgroup Behaviour}
The 2018 test set contained 16,342 false positives and 3,118 false
negatives at the selected threshold. Differences between these groups
were observed in variables including loan amount, interest rate, income, DTI, and revolving utilization. The presence of both error types shows that the model does not perfectly separate the two outcome classes. The subgroup analysis examined employment length, home ownership, loan term, and verification status. Its purpose was to check performance variation across the selected groups rather than to provide a complete fairness evaluation. The study does not include all potentially relevant protected attributes or a full fairness framework. Therefore, the subgroup results should be treated as a robustness check within the available test data.
These findings also support the human-review design of CredWise. Model probabilities should be treated as one source of evidence together with explanations, policy information, and other available information rather than as an unquestionable decision.

\subsection{Retrieval, Analytics, and Agentic Components}
The policy retrieval component achieved Hit@5 = 1.0 for BM25, FAISS,
and the hybrid method on the 28-query labeled evaluation set, with
FAISS reaching the highest MRR (0.964). This shows that the expected
policy section was retrieved for each query in the controlled evaluation.
However, the small evaluation set does not support generalization to
larger or more diverse policy collections \citep{lewis2020rag}.
The SQL benchmark passed all six predefined cases, including expected
query outputs and the tested safety checks. These results validate the
implemented prototype for the tested scenarios, but do not establish
safety for all possible SQL inputs or production databases.
The agent-routing evaluation contained 45 predefined cases covering
decision, policy, risk, and SQL routes and achieved an overall accuracy of 0.956 (43/45 cases; macro F1 = 0.96). The result confirms the behaviour of the tested routing logic but does not establish general agent reliability. Agent systems can introduce routing, tool-use, retrieval, and generation errors, making component-level evaluation important \citep{wang2023agentsurvey,liu2023agentbench}.
The final decision-intelligence evaluation successfully combined model
prediction, SHAP evidence, policy evidence, and analytical context into one structured output. Its role is to organize available evidence rather than produce an independent credit-risk prediction.

\subsection{Human-in-the-Loop Decision Support and Relation to Previous Work}
CredWise is designed so that the system does not make the final lending decision. Instead, it produces a decision-support package for human review. Keeping prediction, retrieved policy evidence, explanations, and analytics as separate evidence sources allows the reviewer to inspect the basis of the generated output. This follows the broader principle of maintaining appropriate human control in AI-assisted workflows \citep{amershi2019guidelines,bussmann2021explainable,bucker2022lending}.
CredWise also builds on earlier work, which studied multi-agent and retrieval-based workflows for enterprise customer support \citep{tiwari2026shopease}. The present work applies related system ideas to credit-risk decision support, with additional emphasis on model explanations, probability calibration, policy evidence, and financial analytics. An earlier study of local LLM judges found that high internal consistency did not necessarily imply agreement with human ratings \citep{tiwari2026consistency}. This motivates the use of predefined component-level tests and human review rather than treating automated agent outputs as automatically reliable.
Overall, the results address the seven research questions by showing
that the final model provides useful temporal test performance, Feature Set B improves the tested predictive metrics, calibration improves probability quality, SHAP rankings remain highly similar across the evaluated periods, and measurable feature drift exists between the training and later periods. The controlled retrieval, SQL, and agent evaluations also produced the expected outputs, while the decision-intelligence layer successfully combined the available
evidence into a human-reviewable output.

\section{Limitations}
\label{limitations}
This study has several limitations that should be considered when
interpreting the results.
The experiments rely on a single historical Lending Club dataset and a
temporal evaluation window ending in 2018 (2007--2016 training, 2017
validation, 2018 test). While this provides a more realistic evaluation than a random split, it does not show how the model would behave under future changes in borrower populations, lending policies, or economic conditions, and the results may not directly transfer to other institutions or portfolios.
The policy retrieval, SQL, and agent-routing components were each
evaluated using small, predefined test sets -- 28 labeled queries for retrieval, six cases for SQL, and 45 cases for agent routing. These benchmarks confirmthe implemented behaviour for the tested
scenarios but are too limited to support claims of general retrieval
accuracy, SQL robustness, or agent reliability under a broader range
of queries and inputs. The decision-intelligence layer was also
evaluated on a single representative applicant, so the result
demonstrates component integration for that case but does not establish robustness across a larger applicant population.
The explainability and drift analyses are descriptive rather than
causal. SHAP values quantify feature contributions to the model output, and PSI quantifies distributional change; neither establishes why a feature is important or why a distribution shifted, and the observed temporal stability of SHAP rankings does not guarantee stability under future data changes.
The subgroup analysis, covering employment length, home ownership,
loan term, and verification status, is a robustness check rather than
a complete fairness evaluation, since it does not cover all
potentially relevant protected or sensitive attributes.
Finally, CredWise is an academic research prototype rather than a
production lending system. It does not provide regulatory, legal, or
financial advice, and the reported experiments do not establish
suitability for autonomous lending decisions; the final decision is
intentionally left to a human reviewer.
Future work can address these limitations through evaluation on
additional datasets and time periods, larger and more diverse
retrieval/SQL/agent benchmarks, and a broader analysis of robustness,
fairness, and deployment-related risks.

\section{Conclusion}
\label{conclusion}
This paper presented CredWise, an academic research prototype for
controlled credit-risk decision support that combines an XGBoost
credit-risk model with probability calibration, SHAP-based
explainability, policy retrieval, SQL analytics, and controlled
agent-based workflows within a single evidence-first architecture.
Using a temporal evaluation design (2007--2016 training, 2017
validation, and 2018 untouched test) over 1,345,310 loan records, the
calibrated model achieved a ROC-AUC of 0.7109 and an F1-score of
0.3714 on the final test period. Sigmoid calibration reduced the
Brier score from 0.2157 to 0.1273 and the expected calibration error
from 0.2862 to 0.0585. These results indicate that the model provides
useful ranking performance and that calibration improves the quality
of the predicted probabilities.
The explanation and drift analyses together highlight an important
observation of this study: model behaviour can remain highly stable
at the feature-importance level, with a Spearman correlation of
0.9959 between the 2017 and 2018 global SHAP rankings, even as the
underlying input distributions shift. The largest observed shifts
were found for \texttt{revol\_util} and \texttt{int\_rate}. This
distinction between explanation stability and data drift suggests
that temporal monitoring can be useful in addition to one-time model
validation.
The supporting components -- policy retrieval, SQL analytics, and
agent routing -- were each evaluated independently against predefined
test cases rather than assessed only through the correctness of a
final generated response. This component-wise evaluation strategy,
combined with the decision-intelligence layer's ability to assemble
model, policy, and analytical evidence into a single structured
output, allows the different sources of evidence to remain traceable
when they are combined for review.
Importantly, CredWise does not automate the lending decision itself.
The system is deliberately constrained to produce decision-support
evidence, with the final judgment left to a human reviewer. This
human-in-the-loop design, together with the separation of model,
policy, and analytical evidence, is central to the auditability goals
of the framework.
The reported results should be interpreted within the scope of a
single historical dataset, a limited temporal window, and small
controlled benchmarks for the retrieval, SQL, and agent components,
as discussed in Section~\ref{limitations}. Broader claims about
real-world deployment, generalization to other lending institutions,
or long-term robustness would require evaluation on additional
datasets, longer time horizons, and larger benchmark sets. Future work along these directions, together with a more comprehensive fairness analysis, would help clarify the conditions under which a system such as CredWise could support, rather than replace, human credit-risk decision-making.

\section*{Data Availability}
The study uses the publicly available Lending Club historical loan
dataset (accepted\_2007\_to\_2018Q4.csv), obtained from Kaggle
(\url{https://www.kaggle.com/datasets/wordsforthewise/lending-club}),
subject to that platform's licensing and redistribution terms. The
CredWise implementation, trained model artifacts, and research
evaluation scripts will be made available in a public repository upon
publication.

\section*{Declaration of Generative AI and AI-assisted Technologies}
During the preparation of this manuscript, the author used ChatGPT
(OpenAI) and Claude (Anthropic) for manuscript organization, language improvement, and
editing. The author reviewed and revised all AI-assisted content,
independently verified the reported results and references, and takes
full responsibility for the final content of the manuscript.

\section*{Acknowledgements}
The author thanks the Indian Institute of Technology Kharagpur for providing the academic environment and computational resources for this research.

\section*{Funding}
This research did not receive any specific grant from funding agencies in the public, commercial, or not-for-profit sectors.

\section*{CRediT authorship contribution statement}
Aakash Kumar Tiwari: Conceptualization, Methodology, Software, Data curation, Formal analysis, Investigation, Visualization, Writing -- original draft, Writing -- review \& editing.

\section*{Declaration of competing interest}
The author declares that he has no known competing financial interests or personal relationships that could have appeared to influence the work reported in this paper.


\bibliographystyle{model5-names}
\biboptions{authoryear}
\bibliography{references}

\end{document}